\documentclass{article}
\usepackage{spconf,graphicx}
\usepackage{xcolor}
\usepackage{stmaryrd}
\usepackage{amsmath,amssymb,amsfonts,amsthm}
\usepackage{optidef}
\usepackage{subcaption}
\usepackage{multirow,booktabs,makecell}

\usepackage{flushend}

\newlength{\bibitemsep}
\newlength{\bibparskip}
\let\oldthebibliography\thebibliography
\renewcommand\thebibliography[1]{%
 \oldthebibliography{#1}%
 \setlength{\parskip}{\bibitemsep}%
 \setlength{\itemsep}{\bibparskip}%
}

\usepackage[all=normal,paragraphs=tight,floats=normal,mathspacing=normal,wordspacing=tight,charwidths=tight,mathdisplays=normal,leading=tight]{savetrees}

\newcommand{\norm}[2]{\ensuremath{\left\lVert #1 \right\rVert}_#2}
\newcommand{\abs}[1]{\ensuremath{\left| #1 \right|}}
\newcommand{\esp}[1]{\ensuremath{\mathbb{E}\left[ #1 \right]}}

\title{Model-based learning for location-to-channel mapping}
\name{Baptiste Chatelier$^{\ddagger,\dagger,\star}$ \qquad Luc Le Magoarou$^{\dagger}$ \qquad Vincent Corlay$^{\ddagger,\star }$ \qquad  Matthieu Crussi\`ere$^{\dagger,\star}$}
  
\address{$^{\dagger}$Univ Rennes, INSA Rennes, CNRS, IETR-UMR 6164, Rennes, France\\
$^{\ddagger}$Mitsubishi Electric R\&D Centre Europe, Rennes\\
$^{\star}$b\raisebox{0.2mm}{\scalebox{0.7}{\textbf{$<>$}}}com, Rennes, France
}
\begin{document}
%
\maketitle
\begin{abstract}

Modern communication systems rely on accurate channel estimation to achieve efficient and reliable transmission of information. As the communication channel response is highly related to the user's location, one can use a neural network to map the user's spatial coordinates to the channel coefficients. However, these latter are rapidly varying as a function of the location, on the order of the wavelength. Classical neural architectures being biased towards learning low frequency functions (spectral bias), such mapping is therefore notably difficult to learn. In order to overcome this limitation, this paper presents a frugal, model-based network that separates the low frequency from the high frequency components of the target mapping function. This yields an hypernetwork architecture where the neural network only learns low frequency sparse coefficients in a dictionary of high frequency components. Simulation results show that the proposed neural network outperforms standard approaches on realistic synthetic data.

\end{abstract}

\section{INTRODUCTION}
\label{sec:intro}

Classical signal processing methods have been used for decades to solve data processing problems. Those methods are model-based and, as every model is imperfect, they potentially exhibit a high bias. However, such methods benefit from a relatively low complexity. Recently, machine learning methods have introduced a paradigm shift: rather than using models, one can use generic neural architectures that learn from data. Such methods exhibit a low bias due to their intrinsic adaptability, but the computational and sample complexity of their training can be very high. Model-based machine learning~\cite{Shlezinger2023} proposes to take the best of both worlds: achieving at the same time low bias and low complexity by using models from signal processing to initialize, structure, and train learning methods.

The field of communication engineering is particularly well adapted to the use of model-based (MB) machine learning as many models have been developed to describe communication systems. More specifically, one can use propagation channel models to achieve accurate channel estimation~\cite{Hengtao18,Xiuhong21,yassine2022,Chatelier2022}. As propagation channels are highly correlated to the user's location, one could learn the location-to-channel mapping whose knowledge could be useful in many applications. Indeed, it makes location-based channel estimation possible, but also radio-environments compression: the radio environment around a base station (BS) would be stored in the weights and biases of the trained neural network. Apart from that, it could also serve beamformer prediction, jamming detection, resource allocation or even secure communication mechanisms. However, the spatial dependence of the underlying model varies on the order of the wavelength, making this mapping remarkably complex to learn. As a matter of fact, it has been proven in~\cite{Rahaman2019} that classical neural architectures tend to be biased towards learning low frequency functions (spectral bias).

\noindent\textbf{Contributions.} In this paper, a physics-based channel model is used to derive a model-based neural architecture aimed at learning the location-to-channel mapping in a supervised manner. Similarly to architectures from the implicit neural representation (INR) literature ~\cite{tancik2020fourfeat,sitzmann2019siren,mildenhall2020nerf} 
, the proposed neural network presents a spectral separation stage that splits the low frequency from the high frequency content of the target mapping function. This allows to bypass the spectral bias issue. However, in opposition to INR architectures, the proposed architecture has an additional hypernetwork whose role is to learn activation coefficients of the high frequency components in a sparse manner. The proposed architecture is evaluated on realistic synthetic data against classical and INR neural architectures. It yields a huge improvement in reconstruction performances but also a drastic reduction in the number of parameters to learn.

\noindent\textbf{Related work.} Learning mappings through neural networks has been extensively studied in the INR community for image reconstruction~\cite{sitzmann2019siren,Bemana2020,Chen2021} and $3$D scene reconstruction from $2$D images~\cite{mildenhall2020nerf,Yariv2020,Sitzmann2019}. A specific focus has been paid on finding architectures that can learn high frequency details~\cite{tancik2020fourfeat,sitzmann2019siren,mildenhall2020nerf}. Moreover, using machine learning methods in order to achieve channel estimation has attracted a huge interest in the past years~\cite{Xuanxuan18,Mehran19,Eren20}. Previous works about the learning of the location or pseudo-location to beamformer mapping exist~\cite{lemagoarou2022,lemagoarou2022_asilomar}. However, to the best of the authors' knowledge, there is no previous work specifically focused on learning the location-to-channel mapping.


\section{PROBLEM FORMULATION}\label{sec:Section_2}
In this paper, a single input single output (SISO) monocarrier scenario with one BS located at $\mathbf{x}_1 = \left(x_1,y_1\right)$ and one user equipment (UE) located at an arbitrary location $\mathbf{x} = \left(x,y\right)$ is considered. Note that $2$D locations are considered here for simplicity of exposition and illustration. However, the proposed method can be straightforwardly extended to the $3$D case. Considering $L_p$ virtual propagation paths, the channel coefficient of the BS-UE link at a given frequency can be modeled as:
\begin{equation}
    h = \sum_{l=1}^{L_p} \gamma_l \mathrm{e}^{-\mathrm{j}\frac{2\pi}{\lambda}d_l},
\end{equation}
where $\gamma_l$ and $d_l$ are the complex attenuation and propagation distance of the $l$th path, and $\lambda$ is the wavelength. Using the image source theory to model the propagation interactions~\cite[Chapter 1, p.47-49]{pozar2011microwave} and expanding the $\gamma_l$ term yields:
\begin{equation}
    h\left(\mathbf{x}\right) = \sum_{l=1}^{L_p} \dfrac{\alpha_l \mathrm{e}^{\mathrm{j} \beta_l}}{\norm{\mathbf{x}-\mathbf{x}_l}{2}} \mathrm{e}^{-\mathrm{j}\frac{2\pi}{\lambda}\norm{\mathbf{x}-\mathbf{x}_l}{2}},\label{eq:spherical}
\end{equation}
where $\forall l>1$, $\mathbf{x}_l \in \mathbb{R}^2$ is the image source location associated to the $l$th path. $\alpha_l$ and $\beta_l$ represent the small-scale attenuation and phase shift of the $l$th path, so that $\alpha_1 = 1$ and $\beta_1 = 0$ when considering a Line of Sight (LoS) path. The $1/\norm{\mathbf{x}-\mathbf{x}_l}{2}$ attenuation represents the large scale fading of the $l$th path.

The goal of this study is to calibrate
\begin{equation}
    f_{\boldsymbol{\theta}}: \mathbf{x} \rightarrow h\left(\mathbf{x}\right),
\end{equation}
a neural network $f$ parameterized by a set of parameters $\boldsymbol{\theta}$ that maps the location $\mathbf{x}$ to its channel coefficient $h\left(\mathbf{x}\right)$. The high frequency spatial dependence of the considered propagation model can be seen in the argument of the exponential in Eq.~\eqref{eq:spherical}. As the carrier frequency rises, the wavelength drops: at usual frequencies used in communication systems (sub-$6$GHz), $\lambda$ is of the order of a few centimeters. Thus, a small variation of the considered location $\mathbf{x}$ leads to a huge change in its channel coefficient $h\left(\mathbf{x}\right)$, making the location-to-channel mapping notably hard to learn.

\section{PROPOSED METHOD}\label{sec:Section_3}
It is well known that deep neural networks are universal function approximators~\cite{Hornik89,Cybenko1989ApproximationBS}. However, it has been shown that classical neural architectures are biased towards learning low frequency functions, a phenomenon known as spectral bias, making them impractical for the learning of rapidly varying functions~\cite{Rahaman2019,ijcai2021p304}. In this paper, it is proposed to take insights from the propagation model to design a neural architecture that learns the location-to-channel mapping, following the model-based machine learning paradigm~\cite{Shlezinger2023}.

\noindent\textbf{Local planar approximation.} The channel model in Eq.~\eqref{eq:spherical} represents waves spherically propagating from the BS antenna and their reflections on various obstacles. Recall that $\mathbf{x}_l \in \mathbb{R}^2$ is the location of the $l$th image source. Around an arbitrary reference location $\mathbf{x}_r \in \mathbb{R}^2$, those spherical wavefronts can be approximated locally by planar ones. This can be shown using a Taylor expansion \cite{Lemagoarou2019}. Let $\xi\left(\mathbf{x}\right) \triangleq \norm{\mathbf{x}-\mathbf{x}_l}{2}$, $\xi\left(\mathbf{x}\right)$ is differentiable at $\mathbf{x} = \mathbf{x}_r$. The first order Taylor expansion of $\xi\left(\mathbf{x}\right)$ around $\mathbf{x}_r$ yields:
\begin{equation}
    \begin{split}
        \xi\left(\mathbf{x}\right) &\simeq \xi\left(\mathbf{x}_r\right) + \left. \nabla \xi \left(\mathbf{x}\right)\right|_{\mathbf{x}_r}\cdot\left(\mathbf{x}-\mathbf{x}_r\right)\\
        &= \norm{\mathbf{x}_r-\mathbf{x}_l}{2} + \mathbf{u}_{\left(\mathbf{x}_r-\mathbf{x}_l\right)}\cdot\left(\mathbf{x}-\mathbf{x}_r\right)\label{eq:local_planar},
    \end{split}
\end{equation}
where $\mathbf{u}_{\left(\mathbf{x}_r-\mathbf{x}_l\right)}$ is the unit norm vector in the $\left(\mathbf{x}_r-\mathbf{x}_l\right)$ direction.

Injecting Eq.~\eqref{eq:local_planar} into Eq.~\eqref{eq:spherical} gives, for $\mathbf{x}$ close to $\mathbf{x}_r$:
\begin{equation}
    h\left(\mathbf{x}\right) \simeq \sum_{l=1}^{L_p} \alpha_l \mathrm{e}^{\mathrm{j}\beta_l} h_l\left(\mathbf{x}_r\right)\dfrac{\mathrm{e}^{-\mathrm{j}\frac{2\pi}{\lambda}\mathbf{u}_{\left(\mathbf{x}_r-\mathbf{x}_l\right)}\cdot\left(\mathbf{x}-\mathbf{x}_r\right)}}{1+\dfrac{\mathbf{u}_{\left(\mathbf{x}_r-\mathbf{x}_l\right)}\cdot \left(\mathbf{x}-\mathbf{x}_r\right)}{\norm{\mathbf{x}_r-\mathbf{x}_l}{2}}},\label{eq:planar_interp}
\end{equation}
where $h_l\left(\mathbf{x}_r\right)$ is the channel coefficient of the $l$th source at the reference location $\mathbf{x}_r$. It is worth noting that the interpolation term, i.e. the fraction, tends to $1$ when $\mathbf{x}_r$ tends to $\mathbf{x}$. Moreover, the numerator of the interpolation term is the equation of a planar wave, propagating in the direction of the vector $\left(\mathbf{x}_r - \mathbf{x}_l\right)$. As a  result, locally, Eq.~\eqref{eq:spherical} can be viewed as a linear combination of planar wavefronts. Such planar wave approximation of spherical waves is at the heart of many array processing techniques, as it gives rise to the well-known steering vectors model, or spatial signature~\cite[Chapter 7]{Tse2005}.

For any location $\mathbf{x} \in \mathbb{R}^2$, Eq.~\eqref{eq:planar_interp} shows that one can approximate the channel coefficient $h\left(\mathbf{x}\right)$ as a linear combination of planar wavefronts. Rearranging the terms, one can rewrite Eq.~\eqref{eq:planar_interp}, separating the high frequency content (planar wavefronts, varying at the wavelength scale) from the low frequency content (coefficients):
\begin{equation}
    h\left(\mathbf{x}\right) \simeq \sum_{l=1}^{L_p} \underbrace{\dfrac{\alpha_l \mathrm{e}^{\mathrm{j}\beta_l} h_l\left(\mathbf{x}_r\right) \mathrm{e}^{\mathrm{j}\mathbf{k}_{r,l}\cdot\mathbf{x}_r}}{1+\dfrac{\mathbf{u}_{\left(\mathbf{x}_r-\mathbf{x}_l\right)}\cdot \left(\mathbf{x}-\mathbf{x}_r\right)}{\norm{\mathbf{x}_r-\mathbf{x}_l}{2}}}}_{\text{Slowly varying}} \underbrace{\mathrm{e}^{-\mathrm{j}\mathbf{k}_{r,l}\cdot\mathbf{x}}\vphantom{\dfrac{\alpha_l \mathrm{e}^{\mathrm{j}\beta_l} h_l\left(\mathbf{x}_r\right) \mathrm{e}^{\mathrm{j}\mathbf{k}_{r,l}\cdot\mathbf{x}_r}}{1+\dfrac{\mathbf{u}_{\left(\mathbf{x}_r-\mathbf{x}_l\right)}\cdot \left(\mathbf{x}-\mathbf{x}_r\right)}{\norm{\mathbf{x}_r-\mathbf{x}_l}{2}}}}}_{\text{Fastly varying}},\label{eq:separation_slow_fast}
\end{equation}
where $\mathbf{k}_{r,l} = \frac{2\pi}{\lambda}\mathbf{u}_{\left(\mathbf{x}_r-\mathbf{x}_l\right)}$ is an angular wave vector in the $\left(\mathbf{x}_r-\mathbf{x}_l\right)$ direction.

As mentioned before, this planar approximation is only valid in a local neighborhood of $\mathbf{x}_r$. Let us tile the space with $N$ hexagons inscribed in circles of center $\mathbf{x}_{r_i}$ and radius such that $\forall \mathbf{x} \in \mathcal{H}_i, \abs{h\left(\mathbf{x}\right)-h^\prime\left(\mathbf{x}\right)} \leq \epsilon$, with $\mathcal{H}_i$ being the location set in the hexagon of center $\mathbf{x}_{r_i}$, and $h^\prime\left(\mathbf{x}\right)$ the Taylor-approximated channel, i.e. the right-hand term of Eq.~\eqref{eq:separation_slow_fast}. This radius is closely related with the spatial-validity length of the Taylor approximation in Eq.~\eqref{eq:local_planar}. Then, Eq.~\eqref{eq:planar_interp} shows that, within each of these hexagons, only $L_p$ planar wavefronts are required to compute $h^\prime\left(\mathbf{x}\right)$. As a result, a dictionary $\mathbf{\Psi}\left(\mathbf{x}\right) = \left\{\psi_i\left(\mathbf{x}\right)\right\}_{i=1}^D$ of size $D \leq L_pN$ containing well-chosen planar wavefronts and an activation vector $\mathbf{w}\left(\mathbf{x}\right) \in \mathbb{C}^D$ can be used as follows to approximate $h(\mathbf{x})$ for any $\mathbf{x} \in \mathbb{R}^2$:
\begin{align}
    &h\left(\mathbf{x}\right)\simeq \sum_{i=1}^{D} w_i\left(\mathbf{x}\right)\psi_i\left(\mathbf{x}\right),\label{eq:dict_approach}\\
    &\text{with} \norm{\mathbf{w}\left(\mathbf{x}\right)}{0} = L_p, \forall \mathbf{x}, \nonumber 
\end{align}
where the dictionary functions are constructed as
\begin{equation}
    \psi_i\left(\mathbf{x}\right) = \mathrm{e}^{-\mathrm{j}\mathbf{k}_i\cdot\mathbf{x}}.
\end{equation}

The directions of the planar wavefronts are defined by the spatial frequencies $\mathbf{k}_i$, i.e. angular wave vector but from a spectral perspective. From Eq.~\eqref{eq:planar_interp} one can observe that the angular wave vector is $\frac{2\pi}{\lambda}\mathbf{u}_{\left(\mathbf{x}_r-\mathbf{x}_l\right)}$, so that $\forall i \in \llbracket 1,D \rrbracket, \norm{\mathbf{k}_i}{2} = \frac{2\pi}{\lambda}$. Due to that norm constraint, the spatial frequencies used to approximate the channel in Eq.~\eqref{eq:separation_slow_fast} all belong to a one-dimensional manifold: the circle of radius $\frac{2\pi}{\lambda}$. This implies that it is possible to build a generic dictionary by sampling the circle while keeping a tight approximation (no curse of dimensionality). Moreover, due to their low frequency nature, the activation coefficients $w_i\left(\mathbf{x}\right)$ should be easily learnable by standard neural network architectures.

\begin{figure}[!h]
    \centering
    \includegraphics[scale=1.2]{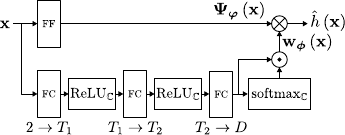}
    \caption{Proposed model-based neural network architecture.}
    \label{fig:mb_net}
\end{figure}

\noindent\textbf{Model-based neural architecture.} In practice, one can use Eq.~\eqref{eq:dict_approach} to derive the model-based neural network presented in Fig.~\ref{fig:mb_net}, in which the dictionary containing high frequency planar wavefronts and the low frequency coefficients are generated from the location in two parallel branches. In more details, the low frequency coefficients in $\mathbf{w}_{\boldsymbol{\phi}}\left(\mathbf{x}\right) \in \mathbb{C}^{D}$ are learned by an hypernetwork~\cite{Schmidhuber1992,Ha2017} of parameter set $\boldsymbol{\phi}$.\footnote{Note that for $z_1 \in \mathbb{C}, \mathbf{z}_2 \in \mathbb{C}^N$, $\texttt{ReLU}_{\mathbb{C}}\left(z_1\right) = \texttt{ReLU}\left(\Re \left(z_1\right)\right) + \mathrm{j} \texttt{ReLU}\left(\Im \left(z_1\right)\right)$ and $\texttt{softmax}_\mathbb{C}\left(\mathbf{z}_2\right) = \texttt{softmax}\left(\abs{\mathbf{z}_2}\right)$.} As integrating the $\ell_0$ constraint of Eq.~\eqref{eq:dict_approach} in the training loss makes it non-differentiable, a solution is to transfer this sparsity constraint on the hypernetwork design. This is done, as seen in Fig.~\ref{fig:mb_net}, with the $\texttt{softmax}_{\mathbb{C}}$ non-linearity which favors sparsity in the activation vector $\mathbf{w}_{\boldsymbol{\phi}}\left(\mathbf{x}\right)$ by greatly attenuating non-significant coefficients. In Fig.~\ref{fig:mb_net}, the Fourier feature (FF) layer, with parameter set $\boldsymbol{\varphi} = \left\{\mathbf{k}_i\right\}_{i=1}^D$, is used to construct the dictionary $\mathbf{\Psi}_{\boldsymbol{\varphi}}\left(\mathbf{x}\right)$, and is defined as:
\begin{equation}
    \text{FF}_{\boldsymbol{\varphi}}:\mathbf{x} \rightarrow \begin{bmatrix}
        \mathrm{e}^{-\mathrm{j} \mathbf{k}_1\cdot\mathbf{x}}, \cdots, \mathrm{e}^{-\mathrm{j} \mathbf{k}_D\cdot\mathbf{x}}
    \end{bmatrix}
\end{equation}
Note that this embedding layer is just the complex interpretation of the $\cos$/$\sin$ embedding in random Fourier features (RFF)~\cite{tancik2020fourfeat,rahimi2007}. Spatial frequencies in the dictionary could be learned, but are kept fixed in this paper, by uniformly sampling the circle of radius $\frac{2\pi}{\lambda}$ (as suggested by the above analysis). 

\vspace{-\baselineskip}

\section{EXPERIMENTS}\label{sec:Section_4}
It is proposed to evaluate the performance of the proposed architecture on synthetic data, with $f=3.5$GHz ($\lambda\simeq 8.5$cm).

\noindent\textbf{Dataset generation.} Firstly, a $10$m by $10$m square scene area is generated. Then the image source locations are defined: a LoS path and static image sources are considered. Constant attenuation coefficients are allocated to each image sources, those coefficients are chosen in a way such that the image sources reflect a large part of the incident wavefronts, typically $0.6<\alpha_i<1, \text{ } i\neq 1$; phase shift values are uniformly sampled between $0$ and $2\pi$. For the training dataset, locations are uniformly sampled in the scene area. A uniform location grid with $\lambda/4$ spacing in both directions is generated for the test dataset, giving around $213$k locations at the selected frequency. Finally, both training and test datasets are generated using Eq.~\eqref{eq:spherical}.

\begin{figure}[!h]
    \centering
    \includegraphics[scale=1.1]{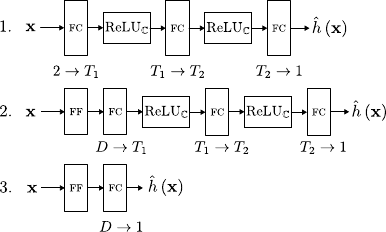}
    \caption{$1.$ MLP; $2.$ RFF; $3.$ RFF lin.}
    \label{figs:net_baseline}
\end{figure}

\noindent\textbf{Models and metrics.} The proposed architecture is compared against a classical MLP and two RFF networks inspired by the INR architectures. They are presented in Fig.~\ref{figs:net_baseline}.

\begin{figure*}[!t]
    \captionsetup{font=small}
    \centering
    \includegraphics[width=2\columnwidth]{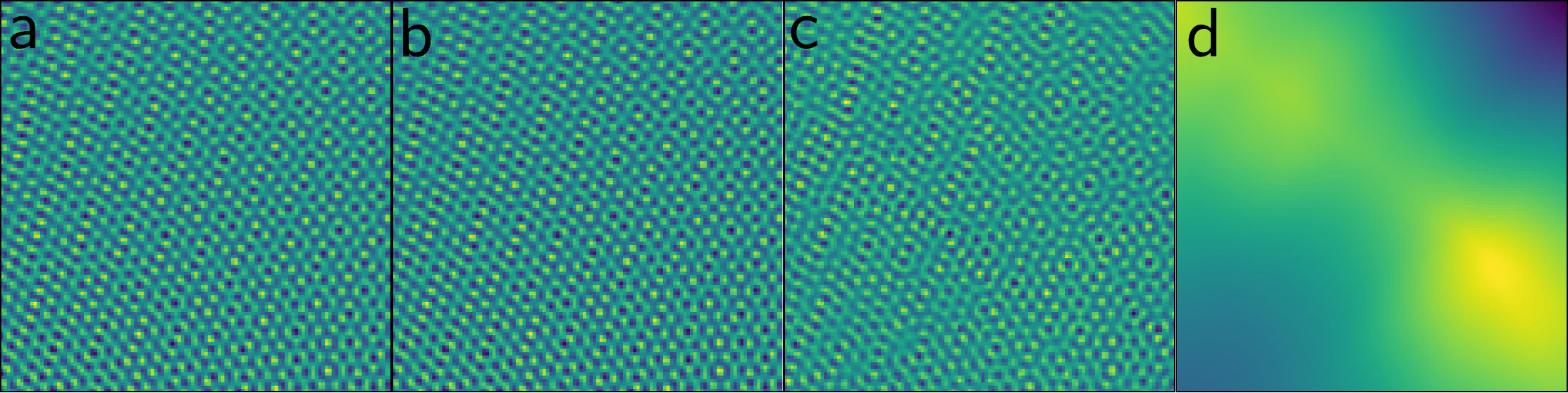}
    \caption{Reconstruction performances (\textit{real part}) over a small zone of the scene area ($2.5$m by $2.5$m), \textsf{a}: Ground truth, \textsf{b}: MB, \textsf{c}: RFF, \textsf{d}: MLP}
    \label{fig:multi}
\end{figure*}

For all baselines, $T_1 = 4096$, $T_2 =2048$ and $D=2000$. In the FF layer, $2.$ and $3.$ use the same uniformly sampled spatial frequencies over the circle of radius $\frac{2\pi}{\lambda}$ than the ones used in the proposed model-based network. For the model-based network, the hyperparameters are fixed as $T_1=256$, $T_2 = 128$ and $D=2000$. The evaluation metric is the Normalized Mean Squared Error (NMSE) in dB over the test dataset, defined as $10\log_{10}(\lVert \mathbf{h}-\hat{\mathbf{h}} \rVert_2^2 / \lVert \mathbf{h} \rVert_2^2)$, where $\mathbf{h}$, resp. $\hat{\mathbf{h}}$, contains the channel coefficients, resp. estimated channel coefficients, over the test dataset. All networks are trained using the $\ell_2$ loss as follows:
\begin{equation}
    \mathcal{L} = \esp{\norm{f_{\boldsymbol{\theta}}\left(\mathbf{x}\right)-h\left(\mathbf{x}\right)}{2}^2}, \mathbf{x} \in \mathcal{D} \subset \mathbb{R}^2,
\end{equation}
where $\mathcal{D}$ is the training location dataset of a particular scene. The complete training dataset is: $\left\{ \mathbf{x}_i, h\left(\mathbf{x}_i\right)\right\}_{i=1}^{N_d}$.

\noindent\textbf{Reconstruction over a specific zone.}
The training location density is equal to $100$ locs./m$^2$ $\simeq 0.7$ locs./$\lambda^2$ which corresponds to $10$k training locations. $L_p = 6$ propagation paths are considered.
\begin{table}[!h]
    \centering
        \begin{tabular}{lccccc}
            \toprule
            &  MLP & RFF & RFF lin. & \textbf{MB}\\
            \midrule
            Params. & $16.8$M & $33.1$M & $4$k & $\mathbf{0.5}$\textbf{M} \\
            \midrule
            $\text{NMSE}_{\text{(dB)}}$ & $0.16$ & $-3.30$ & $-3.04$ & $\mathbf{-20.60}$\\
            \bottomrule
        \end{tabular}
    \caption{NMSE over the test grid.}
    \label{table:nmse_inference}
\end{table}

One can see in Table~\ref{table:nmse_inference} that the proposed model-based architecture outperforms all the baselines, including architectures from the INR literature, while having a much lower parameter complexity than the MLP. Furthermore, it is worth noticing that the proposed model-based network is actually the only one that is able to learn the mapping: the high NMSE values of the other networks show that they fail at this task. One should also remark that the hypernetwork is the key element in the proposed architecture. Indeed, through the model-based initialization of the spatial frequencies, the RFF lin. network output can be seen as a linear combination of planar wavefronts. However, this network fails in the mapping learning task. This also holds true with the over-parametered RFF whose output can be seen as non-linear function of planar wavefronts. In Fig.~\ref{fig:multi} one can see that this network manages to reconstruct high frequency content but still fails in the perfect reconstruction. This is logical as this network possesses high spatial frequencies through its FF layer initialization, but does not fully take advantage of the model analysis. Besides, one can also see in Fig.~\ref{fig:multi} that a model-agnostic MLP also fails to learn the high frequency content of the mapping.

\begin{figure}[!h]
    \centering
    \includegraphics[scale=.5]{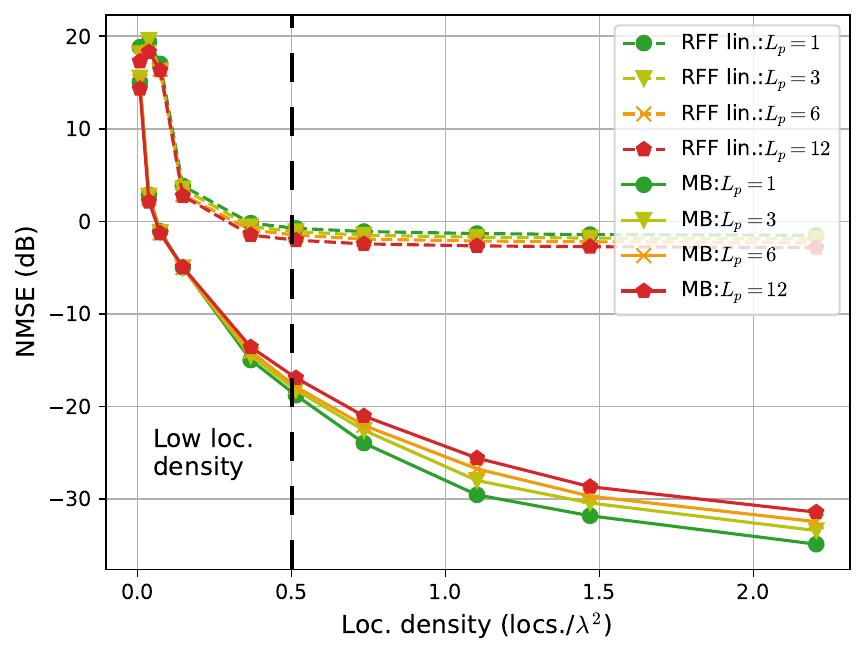}
    \caption{Reconstruction performances}
    \label{fig:lp_dens_var}
\end{figure}

\noindent\textbf{Reconstruction over a specific zone for different $L_p$ and loc. densities, averaged over $100$ trainings.} In this experiment, for each training, new random image source locations are sampled, allowing to simulate multiple scenes. 
Moreover, for each training, all networks are retrained from scratch.
One can see in Fig.~\ref{fig:lp_dens_var} that, for all path and all location density configurations, the model-based network outperforms the baselines. The model-based approach also presents a failure mode for low location density configurations. This can be explained by the lack of spatially close locations in the training dataset in that regime, resulting in the learning failure of the rapidly varying spatial content. The spatial Shannon-Nyquist criterion for perfect reconstruction gives a location density of $4$ locs./$\lambda^2$. The proposed MB network achieves almost perfect reconstruction in sub-Shannon-Nyquist location density, outperforming classical signal processing methods. One should note that, when increasing frequency, the wavelength drops, leading to a higher requirement in the number of training locations to keep the same location density in locs./$\lambda^2$. However, as the frequency rises, the number of propagation paths drops, leading to potentially easier location-to-channel mappings to learn.

\noindent\textbf{Reconstruction over a specific zone generated using ray-tracing.} In this experiment, channels in a $10$m by $10$m square scene area are generated using the Sionna~\cite{sionna} ray-tracing module in Paris, France, with the \textit{Etoile} scenario. As for the other experiments, the training locations are randomly sampled inside the scene area, with a location density of $150$locs./m$^2$ $\simeq 1.1$ locs./$\lambda^2$, which corresponds to $15$k training locations. The test locations are generated along a uniform grid with $\lambda/4$ spacing in both directions. The maximal number of propagation paths inside the scene is $11$, and each path can have at most $3$ consecutive reflections. Diffraction and scattering are not considered here. 
One can see in Table~\ref{table:nmse_inference_sionna} that the MB network also outperforms all baselines on more realistic channels and that it presents NMSE values in the same order of magnitude that the ones in Table~\ref{table:nmse_inference}. 

\begin{table}[!h]
\centering
    \begin{tabular}{lccccc}
        \toprule
        &  MLP & RFF & RFF lin. & \textbf{MB}\\
        \midrule
        Params. & $16.8$M & $33.1$M & $4$k & $\mathbf{0.5}$\textbf{M} \\
        \midrule
        $\text{NMSE}_{\text{(dB)}}$ & $0.14$ & $-2.41$ & $-2.21$ & $\mathbf{-23.41}$\\
        \bottomrule
    \end{tabular}
\caption{NMSE over the test grid (ray-tracing channels).}
\label{table:nmse_inference_sionna}
\end{table}

\section{CONCLUSION}\label{sec:Section_5}
In this paper, a model-based neural architecture was proposed to learn a so-called location-to-channel mapping. The architecture was derived from a propagation model using a local planar approximation yielding a specific network architecture. The performance of the proposed network were evaluated on realistic synthetic data, showing a high performance gain compared to both classical and INR architectures, while having a much lower parameter complexity. Future work will include the refinement of the hypernetwork architecture for better performance in the low location density regime, the optimization of the spatial frequencies' distribution, the consideration of multiple antennas, multiple subcarriers and $3$D locations, which are expected to largely enhance the mapping learning capability.

\vfill\pagebreak


\bibliographystyle{IEEEbib}
\bibliography{refs/refs}

\end{document}